\documentclass{article}

\usepackage[main, final]{neurips_2026}

\usepackage[utf8]{inputenc} 
\usepackage[T1]{fontenc}    
\usepackage{hyperref}       
\usepackage{url}            
\usepackage{booktabs}       
\usepackage{amsfonts}       
\usepackage{nicefrac}       
\usepackage{microtype}      
\usepackage{xcolor}         
\usepackage{amsmath}
\usepackage{algorithm}
\usepackage{algpseudocode}
\usepackage{amsmath}
\usepackage{amsfonts}
\usepackage{amssymb}
\usepackage{multirow}
\usepackage{multicol}
\usepackage{graphicx}
\usepackage{wrapfig}
\usepackage{subcaption}
\usepackage{pifont}
\usepackage{adjustbox}
\usepackage[table]{xcolor}
\title{MESSENGER: \underline{M}emory-\underline{E}nhanced \underline{S}equential \\ \underline{S}cene Flow \underline{E}stimation via Autoregressive \\ \underline{N}ext-Frame Forecastin\underline{G}}

\author{%
 Jiuming~Liu\textsuperscript{1}, \ \
 Jianing~Li\textsuperscript{2}, \ \
  Mengmeng~Liu\textsuperscript{3}, \ \
  Hongyang~He\textsuperscript{4}, \\
  \bf Hesheng Wang\textsuperscript{2}, \ \
  Per Ola Kristensson\textsuperscript{1} \\
   {\textsuperscript{\rm 1} University of Cambridge}\quad
  {\textsuperscript{\rm 2} Shanghai Jiao Tong University}\\
 {\textsuperscript{\rm 3} University of Twente}\quad
 {\textsuperscript{\rm 4} University of Warwick}\\
 \texttt{liujiuming123@gmail.com \quad pok21@cam.ac.uk}
}

\begin{document}

\maketitle

\begin{abstract}
  Scene flow can capture low-level 3D motion displacements in dynamic scenarios. Early pairwise estimators relying on instantaneous two-frame motion lack long-term temporal correlation and also struggle with poor extrapolation ability in future prediction. Although some recent methods attempt to explore multi-frame scene flow estimation in a sequence-to-sequence manner, they typically suffer from heavy computational overhead with increasing input frames and long-horizon prediction degradation due to ineffective motion propagation. To address these problems, we propose a novel memory-enhanced sequential scene flow pipeline, called \textbf{MESSENGER}. To sufficiently mine long-term temporal dependencies naturally within consecutive sequences, a memory buffer is designed by explicitly storing multiple history flow estimates and latent states. For each input frame, the temporally stored flows and states are correlated and retrieved to predict the current initialized flow in a next-frame forecasting manner. Furthermore, we develop an uncertainty-aware reweighting module to filter unreliable retrievals and mitigate accumulated errors. Extensive experiments on nuScenes and Argoverse 2 demonstrate state-of-the-art performance of our MESSENGER, reducing EPE3D by \textbf{71.6\%} on nuScenes and \textbf{67.7\%} on Argoverse 2 in long-horizon future extrapolation. This superiority can be attributed to our designed autoregressive forecasting paradigm, which naturally forces the network to progressively learn the next-frame distribution based on history observations. Code will be released at \url{https://github.com/liujiuming123/Messenger}.
\end{abstract}

\section{Introduction}
Scene flow indicates the 3D dense displacement field across consecutive image or point cloud frames \cite{liu2019flownet3d,wu2020pointpwc,feng2024optimizing,liu2026streamvlo,tang2026binding,dong2026languageconditionedworldmodelingvisual}, which is a fundamental task capturing low-level scene dynamics. Scene flow estimation can be applied in various downstream tasks such as dynamic reconstruction and generation \cite{liu2025topolidm,li2026hy,liu20264dstr,liu2026universe,Deng_2026_CVPR}, motion segmentation \cite{baur2021slim,najibi2022motion,liu2023tracing}, and SLAM systems \cite{zhu2024sni,liu2023translo,deng2026voxelized,liu2024dvlo}.

With the rapid development of deep learning architectures, concurrent methods \cite{wu2020pointpwc,puy2020flot,li2022rigidflow,kittenplon2021flowstep3d} commonly leverage learning-based matching for scene flow estimation in a pairwise paradigm. As a pioneering method, FlowNet3D \cite{liu2019flownet3d} adopts hierarchical feature extraction for iterative flow estimation and refinement. PWCNet \cite{wu2020pointpwc} introduces the Pyramid-Warping-Cost Volume (PWC) structure \cite{sun2018pwc} to estimate flows end-to-end. Subsequent approaches strengthen matching reliability \cite{wang2022matters,cheng2022bi,puy2020flot}, exert rigidity constraints \cite{li2022rigidflow}, or capture robust probabilistic distributions with generative models \cite{liu2024difflow3d,li2025genflow3d}. However, these pairwise-only methods are typically constrained by limited temporal receptive field \cite{zhang2026teflow}, neglecting long-term motion consistency intrinsically within LiDAR sequences, while also struggling with occlusions, motion boundaries, and large displacements \cite{liu2025difflow3d,fu2026egoaction,liu2025dvlo4d}.  

\begin{figure}[t]
    \centering
    \includegraphics[width=1.0\linewidth]{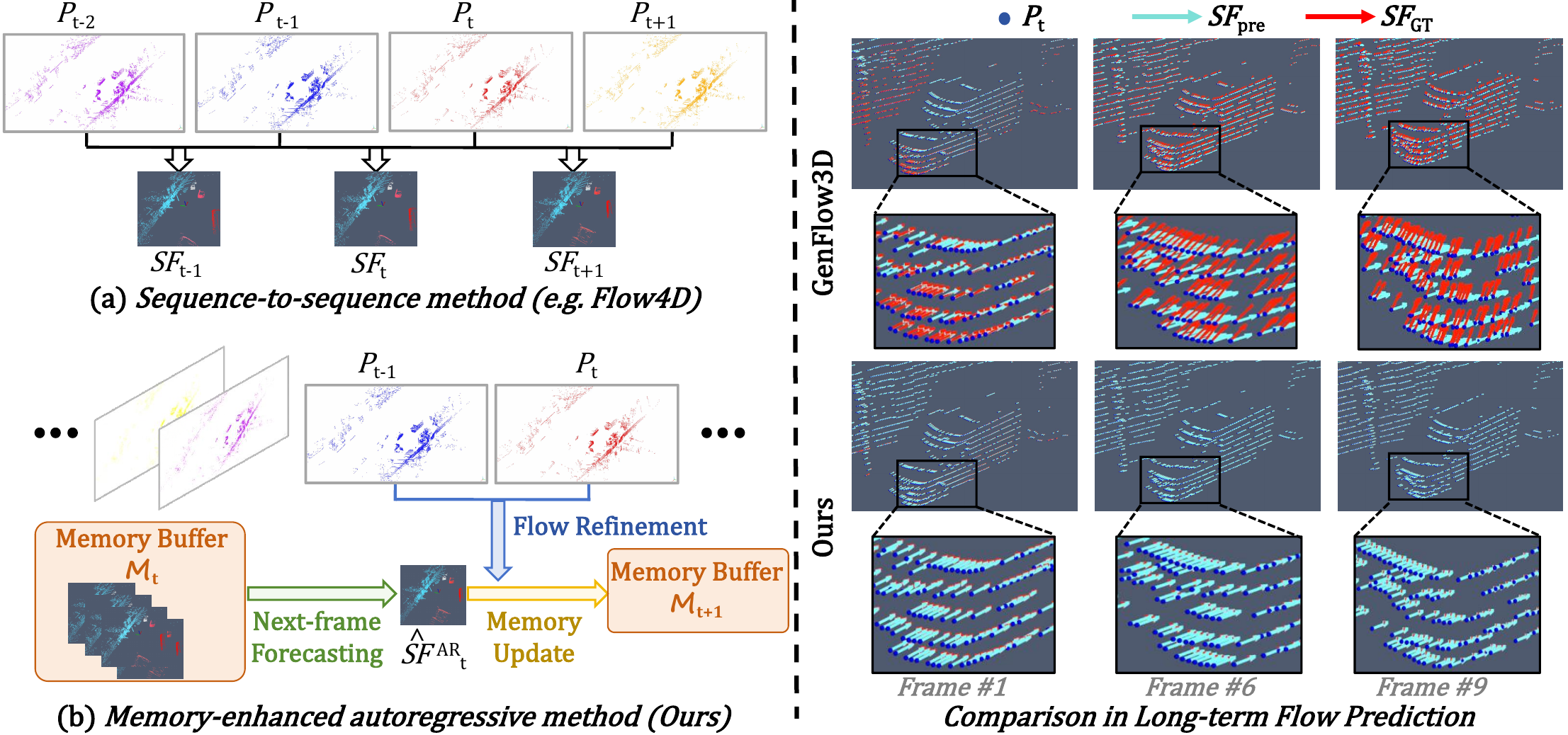}
     \vspace{-0.5cm}
\caption{\textbf{Comparison with previous multi-frame scene flow methods.} (\textbf{\textit{Left}}) Compared to sequence-to-sequence methods with heavy computational costs, our method adopt an autoregressive sequential estimation paradigm, which retrieves multiple history flow priors from a memory buffer. (\textbf{\textit{Right}}) Compared to recent SOTA flow prediction method GenFlow3D \cite{li2025genflow3d} suffering from degraded accuracy in longer-horizon predictions, our uncertainty-aware memory reweighting can mitigate long-term compounding errors by measuring history-present motion discrepancy.
 }
    \label{fig:first}
    \vspace{-0.4cm}
\end{figure}

\begin{wrapfigure}{r}{0.55\textwidth}
  \centering
  \includegraphics[width=0.54\textwidth]{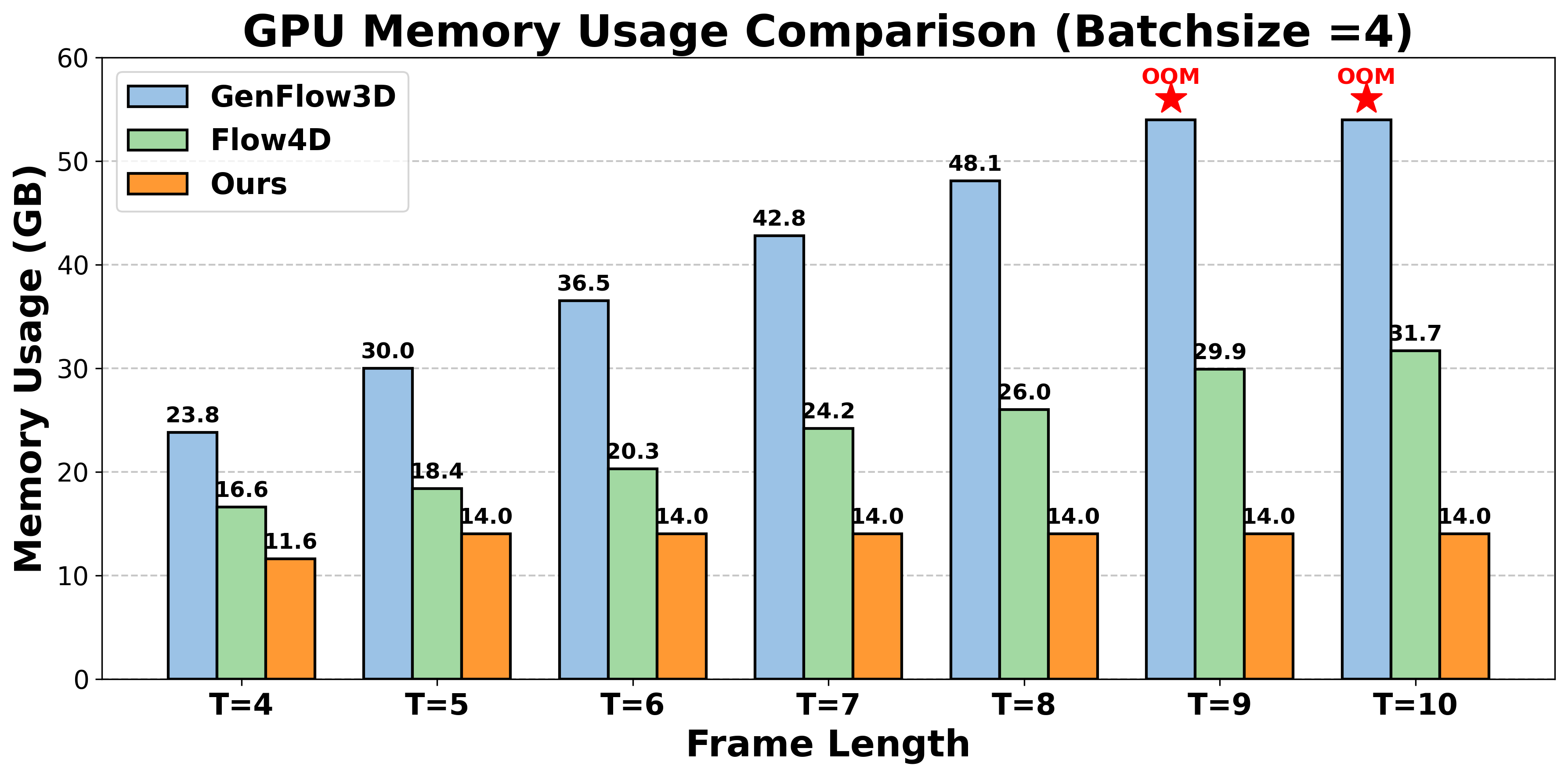}
  \vspace{-10pt}
  \caption{\textbf{GPU memory comparison during training.}}
  \label{fig:gpu}
  \vspace{-10pt}
\end{wrapfigure}

To effectively unlock spatio-temporal dependencies in LiDAR sequences, some researchers recently focus on studying multi-frame scene flow estimation \cite{luo2026mambaflow,kim2025flow4d,zhang2026teflow} as in Fig. \ref{fig:first}. However, sequence-to-sequence methods \cite{kim2025flow4d,luo2026mambaflow,zhang2024deflow} commonly struggle with inevitable increasing computational costs when consuming more frames because they stack multi-frame features in an additional temporal dimension \cite{luo2026mambaflow,kim2025flow4d} or in the channel dimension \cite{zhang2024deflow} in Fig. \ref{fig:first} (\textbf{\textit{Left}}) and Fig. \ref{fig:gpu}. To address this problem, DeltaFlow \cite{zhang2026deltaflow} computes residual flows, rather than full flows, to maintain a constant feature size. Nevertheless, it still requires fixed-length input and also relies on ground truth rigid poses from odometry. Furthermore, previous methods rarely investigate future flow extrapolation potential beyond the observed history contexts. Unlike estimation-only approaches, recent work GenFlow3D \cite{li2025genflow3d} proposes a diffusion-based scene flow estimator that allows future scene flow prediction. However, it faces the challenge of degraded prediction performance along longer-interval extrapolations as in Fig. \ref{fig:first} (\textbf{\textit{Right}}) due to temporally accumulated compounding errors and ineffective implicit motion propagation from the GRU module. Furthermore, the GPU memory usage further restrains its scalability to longer sequences as in Fig. \ref{fig:gpu}.

In this paper, we propose a novel \underline{\textbf{M}}emory-\underline{\textbf{E}}nhanced \underline{\textbf{S}}equential \underline{\textbf{S}}cene flow \underline{\textbf{E}}stimation framework, via autoregressive \underline{\textbf{N}}ext-frame forecastin\underline{\textbf{G}}, called \textbf{MESSENGER}. To sufficiently mine temporal dependencies, a memory buffer is developed to dynamically store explicit coarse flows and implicit latent states from multiple nearest history estimates with minimal computational overhead. When a new frame from LiDAR sequences is delivered, an initial flow is extrapolated by purely correlating and retrieving history memories. Consequently, the forecasting paradigm explicitly enforces the model to predict future motions based on observations. However, the historical retrieval is not equally trustworthy for each point, leading to potential compounding errors during long-horizon predictions. Additionally, it is non-trivial for the forecasting-only flows to consider sudden motion variations and unpredictable dynamics. To alleviate these problems, we also design an uncertainty quantification module to indicate per-point reliability which is used to reweight the memory merging process. In this way, outliers in historical memories that may cause compounding errors will be effectively filtered. Finally, multi-scale diffusion refinement decoders recover fine-grained flow details iteratively. 

Overall, our key contributions in this paper are as follows:
\begin{itemize}
    \item We propose a novel memory-enhanced autoregressive pipeline for joint sequential scene flow estimation and prediction based on the next-frame forecasting mechanism.
    \item A memory buffer stores multi-frame history flows and flow-related hidden states in the coarsest stage that are historically correlated and retrieved to forecast the next-frame flow. This strategy strengthens the long-term temporal consistency while boosting the future flow extrapolation capability.
    \item To rectify erroneous propagation compounding along multiple history memories, we design an uncertainty-aware measurement module to quantify per-point reliability and reweight the memory fusion process.
    \item Our method achieves state-of-the-art performance in both flow estimation and prediction tasks. Notably, our MESSENGER outperforms recent leading method GenFlow3D, reducing EPE3D by 71.6\% on nuScenes and 67.7\% on Argoverse 2 in long-horizon future extrapolation. 
\end{itemize}

\section{Related Work}
\textbf{Pair-wise Scene Flow Estimation.} Early methods utilize stereo \cite{huguet2007variational,li2026combiner,li2026tema} or RGB-D images \cite{hadfield2011kinecting,li2026retrack} as inputs, lacking 3D awareness. As a milestone work, FlowNet3D \cite{liu2019flownet3d} first introduces learnable neural networks for end-to-end flow estimation built on PointNet++ \cite{qi2017pointnet++}. PointPWC-Net \cite{wu2020pointpwc} further introduces the PWC design, which iteratively warps the first frame and calculates cost volume with the second frame. FLOT \cite{puy2020flot} proposes an optimal-transport-based method for point matching. Merging point and voxel representations together, PV-RAFT \cite{wei2021pv} designs both local and global feature extraction. To enhance matching reliability, 3DFlow \cite{wang2022matters}, Bi-PointFlowNet \cite{cheng2022bi}, and MSBRN \cite{cheng2023multi} introduce bi-directional association layers or backward verification modules. To enlarge the receptive field for cross-frame correlation, GMSF \cite{zhang2023gmsf} develops transformer-based scene flow estimation with global matching. Recently, some approaches such as DifFlow3D \cite{liu2024difflow3d,li2025genflow3d} focus on generative probabilistic distributions based on diffusion models. Although these methods have achieved great advancements in the pairwise setting, they cannot capture long-range motion consistency due to limited temporal awareness \cite{liu2026streamvlo}. Additionally, pairwise methods commonly suffer from suboptimal matching performance in challenging scenarios, e.g, occlusion, high dynamics.

\textbf{Multi-Frame Scene Flow Estimation and Prediction.} There is increasing research attention pivoting to multi-frame scene flow estimation enhanced with long-term temporal coherence. Flow4D \cite{kim2025flow4d} proposes a 4D voxel network to extract spatio-temporal information from point cloud sequences. For efficiency, 4D processing is decomposed into 1D and 3D convolutions. MambaFlow \cite{luo2026mambaflow} further reduces computational costs with Mamba backbone \cite{gu2023mamba,liu2024point} achieving linear complexity. To avoid growing features sizes from multiple frames, DeltaFlow \cite{zhang2026deltaflow} develops a $\Delta$ mechanism by computing residual flows rather than directly processing all raw frames. They also design Category-Balanced Loss and Instance Consistency Loss to improve estimation performance. Without requiring labor-intensive ground truth labeling, TEFlow \cite{zhang2026teflow} introduces a temporal ensembling strategy by retrieving flow-related supervisions from a multi-frame candidate pool. However, all these methods can only estimate in-between flows restrained in given observations, which fail to explore the effectiveness of future motion prediction. Also, they mostly rely on preliminary odometry registration results to obtain rigid flow. The most similar paper is GenFlow3D \cite{li2025genflow3d}, which jointly supports estimation and prediction. However, their future prediction performance is extremely limited (e.g., one future frame in their default settings) due to accumulated errors compounding in long-horizon rollout in Fig. \ref{fig:first}.

\textbf{Memory Mechanism.} Memory is widely explored to tackle occluded regions, sensor degradation, and establish temporal dependencies in various computer vision tasks, ranging from semantic segmentation \cite{li2023memoryseg,alonso2021semi}, object detection \cite{chen2020memory,chen2017spatial}, pose estimation \cite{hu2023planning,liu2026regformer++}, etc. The effectiveness of memory has also been proved in optical flow \cite{dong2024memflow,bargatin2025memfof,liu2026arflow}, which facilitates recognition of occluded pixels or motion boundaries. MemFlow \cite{dong2024memflow} designs an attention-based memory read-out and update module combined with a GRU-based refinement layer. MEMFOF \cite{bargatin2025memfof} proposes a memory-efficient RAFT-based flow iteration module, with multi-frame context and correlation features cached and reused in the current cross-frame matching. ARFlow \cite{liu2026arflow} further strengthens the memory construction based on an autoregressive Mamba refinement \cite{liu2025mamba4d,mihajlovic2025pcpmamba}, focusing on geometry-agnostic dynamics across frames. However, we notice that the memory in the corresponding 3D version of optical flow, namely scene flow, is still underexplored. In this paper, we for the first time combine history flows as memory priors for progressive next-frame flow forecasting to learn robust long-term temporal consistency.

\section{Methodology}
\subsection{Preliminaries}
\textbf{Problem Definition.} We propose a novel autoregressive framework on point cloud sequences for \textbf{\textit{joint scene flow estimation and prediction}}. Given $H$-length history observations $\mathcal{P} = \{P_1, P_2, \dots, P_{H}\}$ where $P_t \in N\times 3$, our goal is to estimate the scene flows \textbf{\textit{in between}} $\mathcal{SF}_t (t\leq H)$, representing the per-point displacements from $P_{t-1}$ to $P_{t}$. Due to the explicit sequential memory storage and next-frame forecasting strategy, our method can also extrapolate well to $K$-length \textbf{\textit{future}} scene flow predictions $\mathcal{SF}_t$ ($H<t\leq H+K$) based on historical observations. 

\begin{figure}[t]
    \centering
    \includegraphics[width=1.0\linewidth]{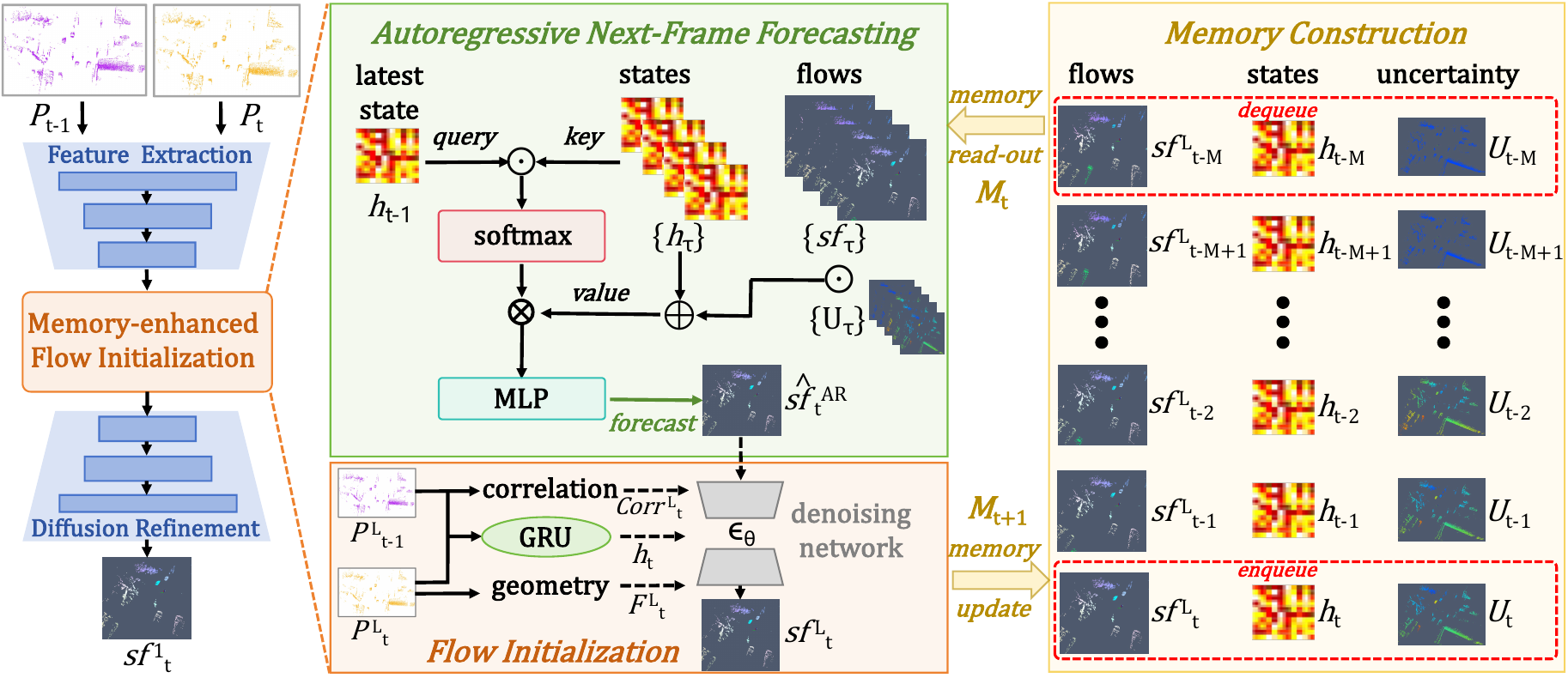}
     \vspace{-0.5cm}
\caption{\textbf{Overall pipeline of our proposed MESSENGER.} 
 }
    \label{fig:pipeline}
    \vspace{-0.4cm}
\end{figure}

\noindent\textbf{Pipeline Overview.} As in Fig. \ref{fig:pipeline}, we adopt a sequential online estimation paradigm where all point sequences are delivered to the network frame-by-frame. For each timestamp $t$, only the current frame $P_{t}$ is additionally inputted. For $P_{t}$, we follow \cite{liu2024difflow3d,li2025genflow3d} to extract $L$-stage hierarchical point features as $\{F_{t}^{l}\}_{l=1}^{L}$. To effectively capture underlying spatio-temporal cues within LiDAR sequences and improve extrapolation capability, we dynamically maintain a memory buffer storing $M$-length temporally nearest history flows. These stored history flows are correlated by a transformer-based memory retrieval module to predict the current coarsest flow in a next-frame forecasting manner as flow initialization. To alleviate the erroneous history fusion and compounding errors, an uncertainty quantification module is designed to filter unreliable history motions. Finally, a multi-scale diffusion decoder \cite{li2025genflow3d} iteratively refines coarse flow in the upper layers.

\subsection{Memory Buffers for Autoregressive Next-Frame Forecasting}
\label{sec:memory}
In real-world LiDAR scans, dynamics mostly tend to be consistent across consecutive frames \cite{zhang2026teflow,dong2024memflow}. Motivated by this observation, we develop a memory buffer $\mathcal{M}$ that explicitly stores multi-frame history coarsest flows to effectively capture long-term spatio-temporal priors. By correlating these history flows, the initial current flow is forecast with enhanced motion propagation. This design offers two advantages: (1) Memory is constructed only in the coarsest layer with \textbf{\textit{minimal computational overhead}}. (2) The progressive next-frame forecasting paradigm naturally learns how to predict future motions from history priors, leading to significantly \textbf{\textit{enhanced flow extrapolation ability}}.

\noindent\textbf{Memory Construction.} At any timestep $t$, the memory buffer maintains a fixed-length queue of historical explicit flows and implicit states from the nearest $M$ frames:
\begin{equation}
    \mathcal{M}_t = \{ (\mathbf{h}_{\tau}, \mathcal{SF}_{\tau}^{L}) \}_{\tau=t-M}^{t-1},
\end{equation}
where $\mathbf{h}_{\tau} \in \mathbb{R}^{n \times C}$ denotes the GRU-based history hidden states following \cite{li2025genflow3d} in the coarsest stage. $\mathcal{SF}_{\tau}^{L} \in \mathbb{R}^{n \times 3}$ is the estimated history flows in the coarsest stage, where $n$ is the point number.


\noindent\textbf{Memory Retrieval for Next-Frame Forecasting.} To retrieve useful motion priors for strengthening the current matching, we formulate the temporal querying process by a cross-attention mechanism. Assuming that motion states are consistent in a short period \cite{li2025genflow3d}, the implicit states can reveal high-level motion tendencies while the latest state retains the highest similarity with the current one. Therefore, We first generate the \textit{query} ($\mathbf{Q}_t$) from the latest flow state and \textit{key} ($\mathbf{K}_{\text{mem}}$) from multiple stacked hidden states:
\begin{equation}
\mathbf{Q}_t = \mathbf{W}_Q \mathbf{h}_{t-1}, \quad
\mathbf{K}_{\text{mem}} = [\mathbf{W}_K \mathbf{h}_{t-M} \oplus \dots \oplus \mathbf{W}_K \mathbf{h}_{t-1}],
\end{equation}

where $\oplus$ indicates feature concatenation. Crucially, the \textit{value} ($\mathbf{V}_{\text{mem}}$) matrix should consider both low-level explicit historical flows and high-level implicit motion tendencies as:
\begin{equation}
\mathbf{V}_{\text{mem}} = \mathbf{W}_V [[\mathbf{h}_{t-M} \oplus \mathcal{SF}^{L}_{t-M}] \oplus \dots \oplus [\mathbf{h}_{t-1} \oplus \mathcal{SF}^{L}_{t-1}]],
\end{equation}
where $\mathbf{W}_Q, \mathbf{W}_K$, and $\mathbf{W}_V$ are learnable linear projections. Afterwards, by computing the scaled dot-product attention, the network aggregates \textbf{\textit{historical}} flows and states from the memory buffer, generating the predicted \textbf{\textit{current-timestamp}} flow $\hat{\mathcal{SF}}_t^{\text{AR}}$ in a next-frame forecasting paradigm:
\begin{equation}\hat{\mathcal{SF}}_t^{\text{AR}} = \text{MLP}\left(\text{Softmax}\left(\frac{\mathbf{Q}_t \mathbf{K}_{\text{mem}}^\top}{\sqrt{d_k}}\right) \mathbf{V}_{\text{mem}}\right),
\end{equation}
where $d_k$ is the feature dimension in transformer. The next-frame forecasting can guarantee long-horizon motion consistency and smooth trajectories by attending to past flow priors. 

\noindent\textbf{Memory Update}. The coarse flow directly predicted from history above lacks current-frame geometry details and motion variations. Therefore, we also refine it with a diffusion-based initialization module \cite{liu2025difflow3d} conditioned by current geometric features $F_{t}^{L}$, correlation features \cite{wu2020pointpwc} $Corr_{t}^{L} = \text{CostVolume}(F_{t-1}^{L},F_{t}^{L})$, and current hidden states \cite{li2025genflow3d} $h_{t} = \text{GRU}(h_{t-1},F_{t}^{L},Corr_{t}^{L})$. The refined flow $\mathcal{SF}_t^{L}$ is then stored to dynamically update our memory buffer in a First-In-First-Out (FIFO) manner in Fig. \ref{fig:pipeline}. The newest generated couple $(\mathbf{h}_t, \mathcal{SF}_t^{L})$ is pushed into $\mathcal{M}_{t+1}$, while the farthest-frame memory $(\mathbf{h}_{t-M}, \mathcal{SF}_{t-M}^{L})$ is discarded for the next-frame ($P_{t+1}$) flow estimation.

\subsection{Uncertainty-Guided Memory Reweighting} 
Forecasting the current flow from history memories fails to consider cross-frame chronological inconsistency or even introduce outliers, especially when objects undergo \textbf{\textit{sudden non-rigid deformations}}, \textbf{\textit{abrupt acceleration}}, or \textbf{\textit{prolonged occlusions}} \cite{liu2023regformer}. Furthermore, the autoregressive paradigm commonly suffers from \textbf{\textit{compounding errors}}: erroneous predictions from previous time steps accumulate and propagate to later frames \cite{chen2024diffusion}. To prevent our progressive flow estimation from being misled by flawed historical estimates, we propose a learnable uncertainty quantification map to dynamically reweight per-frame feature reliability during memory retrieval.

\textbf{Discrepancy-based Uncertainty Measurement.} We hypothesize that the uncertainty of historical memory can be quantified by the discrepancy between temporal motion priors and currently-refined dynamics. We compute the historical-current flow divergence with a flow residual as:
\begin{equation}
\mathbf{Res}_t = \hat{\mathcal{SF}}_t^{\text{AR}} - {\mathcal{SF}}_t^{L},
\end{equation}
where $\mathbf{Res}_t \in \mathbb{R}^{n \times 3}$ is per-point corresponding to the current point cloud $P_{t}$. A large magnitude of $\mathbf{Res}_t$ indicates that the historically interacted flow $\hat{\mathcal{SF}}_t^{\text{AR}}$ strongly contradicts the current geometric evidence, which reveals potential tracking failure or unpredictable inconsistent motions. We then map the residual to a continuous uncertainty scalar $\mathbf{U}_t^{(i)} \in [0, 1]$ for each point $i$ via a L2-Normalization followed by a clamping function:
\begin{equation}
\mathbf{U}_t^{(i)} = \min\left(\max\left(\text{Norm}\|\mathbf{Res}_t^{(i)}\|_{2}, 0\right), 1\right).
\end{equation}

\textbf{Uncertainty-Guided Memory Reweighting.} Quantified uncertainty is then used to dynamically gate the history flows retrieved from the memory buffer, filtering unreliable outliers and disabling inconsistent motion propagation. Specifically, multi-frame reliability $\mathbf{\Gamma}_{\tau} = \mathbf{1} - \mathbf{U}_{\tau}$ is also stored in the memory buffer (Section \ref{sec:memory}) as memory triplets: $(\mathbf{h}_{\tau}, \mathcal{SF}_{\tau}^{L}, \mathbf{\Gamma}_{\tau})$. During the memory retrieval, historical flows are reweighted by this reliability factor per-point, obtaining uncertainty-aware \textit{value}: 
\begin{equation}
\mathbf{V}_{\tau} = \mathbf{W}_V [\mathbf{h}_{\tau} \oplus (\mathcal{SF}_{\tau}^{L} \odot (\mathbf{1} - \mathbf{U}_{\tau}))].
\end{equation}
Through reliability-guided reweighting, the network adaptively penalizes the influence of highly uncertain memory (e.g., with occlusion or dynamics). Consequently, it forces the model to rely more heavily on the current correlation when the history is deemed untrustworthy, while still exploiting robust historical priors to smooth trajectories in confident regions. Finally, multiple cascaded diffusion refinement decoders \cite{liu2025difflow3d} iteratively upsample and refine coarse flow estimates with more fine-grained details, yielding multi-scale scene flow estimates $\{\mathcal{SF}_{t}^{l}\}_{l=1}^{L-1}$.

\begin{table*}[t]
\centering
\caption{\textbf{Comparison of scene flow estimation results on nuScenes \cite{caesar2020nuscenes} and Argoverse 2 \cite{wilson2023argoverse}}. We follow \cite{li2025genflow3d}, inputting $H=3$ consecutive frames to estimate two corresponding flows in between.}
\vspace{-2mm}
\label{tab:comparison}
\resizebox{\textwidth}{!}{
\begin{tabular}{l|c|c|ccc|ccc}
\hline\toprule
\multirow{3}{*}{Method}& \multirow{3}{*}{Reference}& \multirow{3}{*}{Dataset} & \multicolumn{6}{c}{Scene Flow Estimation}  \\ \cline{4-9} \noalign{\smallskip}
 & & & \multicolumn{3}{c|}{$\mathcal{SF}_1$} & \multicolumn{3}{c}{$\mathcal{SF}_2$}  \\ \cline{4-9} \noalign{\smallskip}
 & & & EPE3D $\downarrow$ & Acc3DS $\uparrow$ & Out3D $\downarrow$ & EPE3D $\downarrow$ & Acc3DS $\uparrow$ & Out3D $\downarrow$  \\ \hline \noalign{\smallskip}
PointPWC-Net \cite{wu2020pointpwc} &ECCV'20& \multirow{7}{*}{nuScenes} & 0.0386 & 0.7668 & 0.6255 & 0.0386 & 0.7695 & 0.6212 \\
FLOT \cite{puy2020flot} &ECCV'20 &  & 0.0297 & 0.8561 & 0.5071 & 0.0293 & 0.8604 & 0.5049 \\
PV-RAFT \cite{wei2021pv}  &CVPR'21&  & 0.0500 & 0.7570 & 0.6055 & 0.0492 & 0.7598 & 0.6024  \\
Bi-PointFlowNet \cite{cheng2022bi}  &ECCV'22&  & 0.0291 & 0.8676 & 0.4924 & 0.0291 & 0.8696 & 0.4913\\
PT-FlowNet \cite{fu2023pt}  &RA-L'23&  & 0.0264 & 0.8776 & 0.4753 & 0.0260 & 0.8821 & 0.4695 \\
DiffFlow3D \cite{liu2024difflow3d}  &CVPR'24&  & {0.0222} & {0.9205} & {0.4132} & {0.0221} & {0.9216} & {0.4118} \\
GenFlow3D \cite{li2025genflow3d}  &ICCV'25&  & \underline{0.0161} & \underline{0.9580} & \underline{0.3399} & \underline{0.0138} & \underline{0.9756} & \underline{0.3071} \\ 
\textbf{Ours} & ---&&\textbf{0.0154} &\textbf{0.9629} &\textbf{0.3283}&\textbf{0.0133} &\textbf{0.9767}& \textbf{0.2973} \\
\midrule
PointPWC-Net \cite{wu2020pointpwc}  &ECCV'20& \multirow{7}{*}{Argoverse 2} & 0.0553 & 0.6726 & 0.3681 & 0.0556 & 0.6742 & 0.3682  \\
FLOT \cite{puy2020flot}  &ECCV'20&  & 0.0489 & 0.7681 & 0.3545 & 0.0491 & 0.7678 & 0.3569 \\
PV-RAFT \cite{wei2021pv}  &CVPR'21&  & 0.0968 & 0.6588 & 0.4599 & 0.0971 & 0.6626 & 0.4587 \\
Bi-PointFlowNet \cite{cheng2022bi}  &ECCV'22&  & 0.0298 & 0.8995 & 0.2522 & 0.0301 & 0.8980 & 0.2543 \\
PT-FlowNet \cite{fu2023pt}  &RA-L'23&  & 0.0386 & 0.8161 & 0.3163 & 0.0391 & 0.8154 & 0.3195 \\
DiffFlow3D \cite{liu2024difflow3d}  &CVPR'24&  & {0.0190} & {0.9635} & {0.1872} &{0.0189} & {0.9627} & {0.1887}  \\
GenFlow3D \cite{li2025genflow3d}  &ICCV'25&  & \underline{0.0175} & \underline{0.9660} & \underline{0.1823} & \underline{0.0164} & \underline{0.9711} & \underline{0.1771} \\ 
\textbf{Ours} & ---&&\textbf{0.0167} &\textbf{0.9691} &\textbf{0.1760}&\textbf{0.0156} &\textbf{0.9743}& \textbf{0.1712} \\
\bottomrule\hline
\end{tabular}
}
\end{table*}

\subsection{Loss Functions}
Following previous diffusion-based scene flow methods \cite{li2025genflow3d,liu2025difflow3d}, we adopt multi-scale supervised loss functions on both estimated flows and flow residuals. Specifically, for each point frame $t$ at each stage $l$, two loss functions are formulated as:
\begin{equation}
    L_{res,t}^{l} = || \Delta\overline{\mathcal{SF}}_{t}^{l}-\Delta\mathcal{SF}_{t}^{l}||, \quad
    L_{flow,t}^{l} = ||\overline{\mathcal{SF}}_{t}^{l}-\mathcal{SF}_{t}^{l}||,
\end{equation}
where $\Delta\overline{\mathcal{SF}}_{t}^{l}$ and $\overline{\mathcal{SF}}_{t}^{l}$ denote the ground truth flow residual and full flow. For the layer $l$ from $L-1$ to 1, full flows are recovered by adding estimated flow residuals derived from diffusion refinement onto upsampled coarse flows \cite{liu2024difflow3d}: $\mathcal{SF}_{t}^{l}=\Delta\mathcal{SF}_{t}^{l}+{up}(\mathcal{SF}_{t}^{l+1})$. During training, both $H$ observed frames and $K$ future predicted frames are calculated:
\begin{equation}
    L_{total}=\sum_{t=1}^{H+K-1}\sum_{l=1}^{L}\alpha^{l}(L_{res,t}^{l}+L_{flow,t}^{l}),
\end{equation}
where $\{\alpha^{l}\}_{l=1}^{L}$ are weights in different stages.

\section{Experiments}
\subsection{Datasets and Implementation Details}
\textbf{Evaluation Datasets.} We leverage real-world LiDAR sequences in nuScenes \cite{caesar2020nuscenes} and Argoverse 2 \cite{wilson2023argoverse} for evaluation. The nuScenes dataset captures LiDAR points in a 20 HZ frequency and chooses keyframes with 10-frame intervals. We follow \cite{li2025genflow3d} to construct evaluation sequences, using 650 frames for training and the other 200 frames for testing. For the Argoverse 2 dataset, it scans in a 10 HZ frequency. Following \cite{li2025genflow3d}, we curate training (700 scenes) and evaluation (150 scenes) subsequences, and exclude invalid points into loss computation. Odometry and tracking results are used to calculate ground truth labels. We follow \cite{liu2019flownet3d,puy2020flot,wei2021pv}, choosing EPE3D, Acc3DS, and Out3D as our main metrics, representing average estimation errors, accuracy, and outlier rate, respectively.

\noindent\textbf{Implementation Details.} Unlike GenFlow3D \cite{li2025genflow3d} which can only process short fixed-length sequences, our autoregressive formulation can support arbitrary-length sequential input. Thus, we train our model on all LiDAR frames within the same sequence. When testing, for scene flow estimation, we set the existing length as $H=3$. For scene flow prediction, we set the existing length as $H=6$ as default and varies the future frame length $K$ from 1 to 4. Following \cite{wu2020pointpwc,puy2020flot,gu2019hplflownet,liu2025difflow3d}, 8192 points are randomly sampled, and ground points and points farther than 35m are filtered. The learning rate is initialized as 0.001
and reduced by 0.7 every 40 epochs until convergence. We adopt the AdamW optimizer with $\beta_1 = 0.9, \beta_2 = 0.999$. The diffusion decoder has $L=4$ stages and their loss weights are set as $\{\alpha^{l}\}_{l=1}^{4} = \{0.02, 0.04, 0.08, 0.16\}$, respectively. We leverage DDIM \cite{song2020denoising} with the total timestamp of 1000 while using 1-step denoising during the inference following \cite{liu2024difflow3d}. The length of our designed memory buffer is $M=6$.

\begin{figure}[t]
    \centering
    \includegraphics[width=1.0\linewidth]{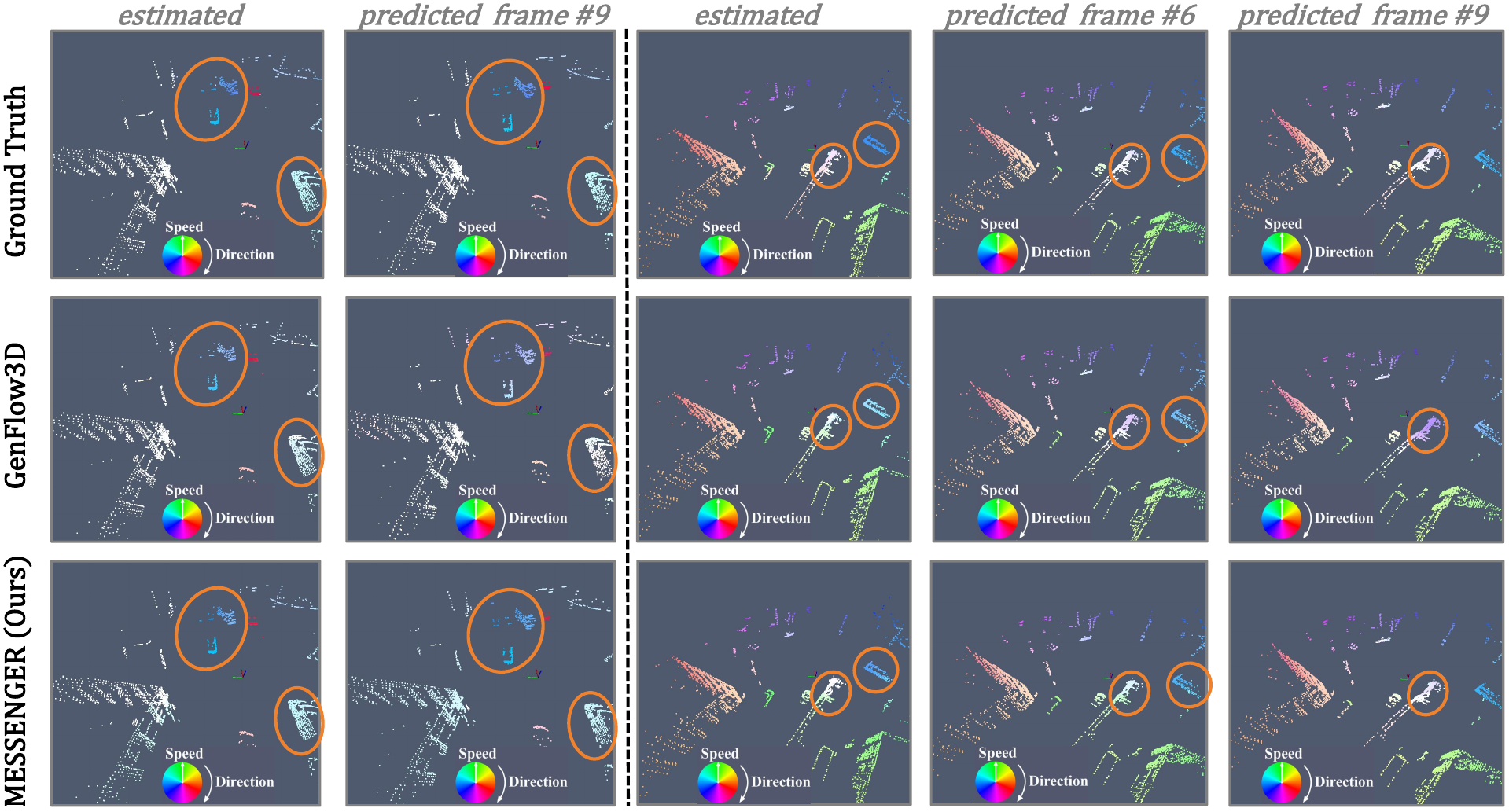}
     \vspace{-0.5cm}
\caption{\textbf{Qualitative comparison in scene flow estimation and prediction on nuScenes.} From up to bottom, rows show ground truth, GenFlow3D \cite{li2025genflow3d}, and our predictions across time. Scene flow is visualized with hue indicating direction and saturation representing speed. Comparison is highlighted with orange circles. Please zoom in for details.
 }
    \label{fig:visual1}
    \vspace{-0.4cm}
\end{figure}

\begin{table}[t]
\centering
\caption{\textbf{Comparison of scene flow prediction results on nuScenes \cite{caesar2020nuscenes} and Argoverse 2 \cite{wilson2023argoverse}}. We set the observed frames as $H$ = 6 and predict future frames with the intervals of $k=1, 2, 3, 4$.}
\label{tab:comparison2}
\resizebox{\textwidth}{!}{
\begin{tabular}{ll|cc|cc|cc|cc}
\hline\toprule
\multirow{4}{*}{\begin{tabular}[c]{@{}c@{}}\rotatebox{90}{nuScenes}\end{tabular}}&\multirow{2}{*}{Methods}& \multicolumn{2}{c|}{$\mathcal{SF}_{H+1}$} & \multicolumn{2}{c|}{$\mathcal{SF}_{H+2}$} & \multicolumn{2}{c|}{$\mathcal{SF}_{H+3}$}& \multicolumn{2}{c}{$\mathcal{SF}_{H+4}$} \\ \cline{3-10} \noalign{\smallskip}
 && EPE3D $\downarrow$ & Acc3DS $\uparrow$& EPE3D $\downarrow$ & Acc3DS $\uparrow$& EPE3D $\downarrow$ & Acc3DS $\uparrow$& EPE3D $\downarrow$ & Acc3DS $\uparrow$ \\ \cline{2-10} \noalign{\smallskip}

&GenFlow3D \cite{li2025genflow3d}  & 0.029 & 0.948 & 0.046 & 0.889 & 0.065 & 0.830 &0.088& 0.754\\ 
&\textbf{Ours} &\bf 0.015&\bf0.969&\bf0.018 &\textbf{0.938} &\textbf{0.021}&\textbf{0.915} &\textbf{0.025}& \textbf{0.883} \\ \midrule
\multirow{4}{*}{\begin{tabular}[c]{@{}c@{}}\rotatebox{90}{Argoverse 2}\end{tabular}}&\multirow{2}{*}{Methods}& \multicolumn{2}{c|}{$\mathcal{SF}_{H+1}$} & \multicolumn{2}{c|}{$\mathcal{SF}_{H+2}$} & \multicolumn{2}{c|}{$\mathcal{SF}_{H+3}$}& \multicolumn{2}{c}{$\mathcal{SF}_{H+4}$} \\ \cline{3-10} \noalign{\smallskip}
 && EPE3D $\downarrow$ & Acc3DS $\uparrow$& EPE3D $\downarrow$ & Acc3DS $\uparrow$& EPE3D $\downarrow$ & Acc3DS $\uparrow$& EPE3D $\downarrow$ & Acc3DS $\uparrow$ \\ \cline{2-10} \noalign{\smallskip}

&GenFlow3D \cite{li2025genflow3d}  & 0.049 & 0.651 & 0.105 & 0.279 & 0.140 & 0.172 & 0.192& 0.130\\ 
&\textbf{Ours} & \bf 0.025&\bf 0.915&\textbf{0.046} &\textbf{0.713} &\textbf{0.054}&\textbf{0.636} &\textbf{0.062}& \textbf{0.555} \\
\bottomrule\hline
\end{tabular}
}
\vspace{-0.4cm}
\end{table}

\subsection{Comparison in Scene Flow Estimation}
\begin{figure}[t]
    \centering
    \includegraphics[width=1.0\linewidth]{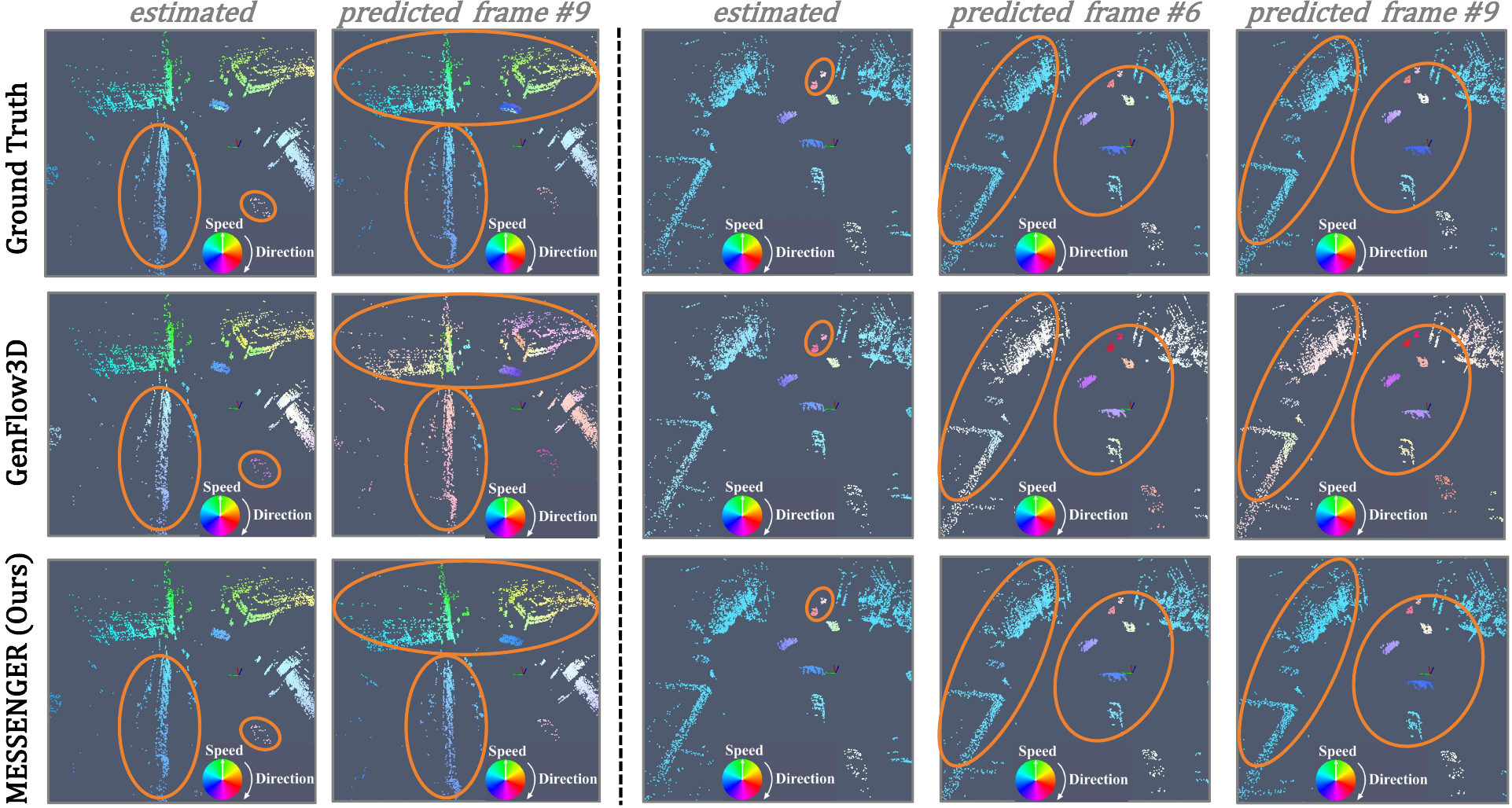}
     \vspace{-0.5cm}
\caption{\textbf{Qualitative comparison in scene flow estimation and prediction on Argoverse 2.} From up to bottom, rows show ground truth, GenFlow3D \cite{li2025genflow3d}, and our predictions across time. Scene flow is visualized with hue indicating direction and saturation representing speed. Comparison is highlighted with orange circles. Please zoom in for details.
 }
    \label{fig:visual2}
    \vspace{-0.2cm}
\end{figure}

To assess the effectiveness of our proposed scene flow estimator, we follow \cite{li2025genflow3d} to deliver three frames as input and estimate two corresponding flows in between on both nuScenes and Argoverse 2. As in Table \ref{tab:comparison}, our method outperforms all previous scene flow estimators in terms of all evaluation metrics, including EPE3D, Acc3DS, Out3D. Compared to the recent SOTA approach GenFlow3D \cite{li2025genflow3d}, our MESSENGER has lower EPE3D and Out3D metrics, and also achieves higher Acc3DS. This advantage is further demonstrated in Fig. \ref{fig:visual1} and Fig. \ref{fig:visual2}, where our estimated flows are more accurate than GenFlow3D's output, especially in challenging scenarios such as dynamic cars and repetitive patterns. However, we notice that the discrepancy between GenFlow3D and ours is small in 3-frame input, mainly because the temporal information is limited. Therefore, we further increase the input frames in Fig. \ref{fig:visual3}, where GenFlow3D has triple higher errors, while ours has even lower errors. 

\subsection{Comparison in Future Scene Flow Prediction}
We also compare the extrapolation ability of future scene flow prediction in Table \ref{tab:comparison2}. We input $H=6$ observed frames and then predict the future flows with a length of $K=4$. On both nuScenes and Argoverse 2, our method has superior extrapolation ability, especially in much longer predictions. When forecasting one future frame, our method has only half of the EPE3D estimation error of GenFlow3D \cite{li2025genflow3d}. The performance gap is further enlarged with the extrapolation frame getting longer from $k=1$ to 4. As for the 4-th frame forecasting, our method has less than 1/3 estimation errors compared to GenFlow3D. On Argoverse dataset, GenFlow3D witnesses obvious decreasing Acc3DS from 0.651 to 0.130. In contrast, ours has way lower drop from 0.915 to 0.555. The significantly improved extrapolation capability is attributed to our designed autoregressive next-frame forecasting module, which naturally learns how to predict future flows based on history observed ones. Furthermore, our uncertainty-aware weighting can filter outliers in historically stored flows, ensuring high temporal consistency. In Fig. \ref{fig:visual1} and Fig. \ref{fig:visual2}, we visualize the long-term prediction results. For the 6-th frame prediction and the 9-th frame, GenFlow3D witnesses huge performance degradation with much larger drifts on both foreground and background objects. Nevertheless, our method still keeps consistently accurate predictions with much longer rollouts.

\subsection{Efficiency Analysis}

\begin{wraptable}{r}{0.55\textwidth}
\centering
 \vspace{-13pt}
\caption{\textbf{Latency comparison.}}
\vspace{-3mm}
\resizebox{1\linewidth}{!}{ 
\begin{tabular}{lccc}
\hline\toprule
Methods&DifFlow3D \cite{liu2024difflow3d}&GenFlow3D \cite{li2025genflow3d}&Ours\\
\midrule
Latency&216.7ms&88.7ms&55.9ms\\
\bottomrule\hline
\end{tabular}
}
\label{tab:latency}
\vspace{-12pt}
\end{wraptable}
In Fig. \ref{fig:gpu}, we have already compared the GPU memory where our GPU consumption is constant to the input frame length. Here, we also additionally compare the latency with previous diffusion-based methods DifFlow3D and GenFlow3D. As in Table \ref{tab:latency}, compared to the other two methods, our method achieves the highest efficiency in terms of one forward latency.

\subsection{Ablation Studies}
\textbf{Memory Construction.} In Table \ref{tab:ablation_arch}, we remove the designed memory buffer, leading to more than triple increasing errors. This demonstrates the extrapolation ability relies significantly on explicit memory storage with next-frame prediction. We also conduct ablations with various memory length in Table \ref{tab:ablation_length}, where from $M=3$ to 6, the prediction quality becomes better. However, storing more frames leads to a slight performance degradation partly because farther frames share scarce semantic similarities with the current one, which can be viewed as outliers. We also compare various storage contents in memory buffers as in Table \ref{tab:ablation_memory}. Comparing a) and b) with e), both explicit flow storage and implicit state storage are vital to the prediction performance. 

\noindent\textbf{Uncertainty.} Without designed uncertainty, there is almost double EPE3D increase in Table \ref{tab:ablation_arch}. This proves that uncertainty can effectively rectify long-term error accumulations. We also compare different construction methods where Normalization-based manner is better than the MLP-based one. In memory buffer, we explicitly store prior-frame uncertainty estimates. Without storing uncertainty in Table \ref{tab:ablation_memory}, there is obviously worse prediction comparing c) to a), and d) to b).

\begin{table*}[t]
  \caption{\textbf{Ablation studies on the nuScenes dataset.} We evaluate the impact of core architecture designs, memory lengths, and memory stored contents in the 4-th future prediction setting.}
  \vspace{-1mm}
  \label{tab:ablation}
  \centering
  \footnotesize
  \setlength{\tabcolsep}{3pt}
  \renewcommand{\arraystretch}{1.2}
  \setlength{\tabcolsep}{1.4mm}
  \begin{subtable}[t]{0.31\linewidth}
      \centering
      \caption{Core novelty and uncertainty.}
      \vspace{-1mm}
      \label{tab:ablation_arch}
      \adjustbox{max width=\linewidth}{
      \begin{tabular}{l|c|c}
        \hline\toprule
       & EPE3D $\downarrow$ & Acc3DS $\uparrow$ \\
        \midrule
        w/o Memory & 0.088&0.754\\
        w/o Uncertainty &0.045 &0.668\\
        MLP-based Uncertainty &0.029 & 0.844\\
         \cellcolor{gray!20}Norm-based Uncertainty &\cellcolor{gray!20}\bf 0.025&\cellcolor{gray!20}\bf 0.883\\
        \bottomrule\hline
      \end{tabular}
      }
  \end{subtable}
  \hfill
  \setlength{\tabcolsep}{5.5mm}
  \begin{subtable}[t]{0.31\linewidth}
      \centering
      \caption{Memory lengths.}
      \vspace{-1mm}
      \label{tab:ablation_length}
      \adjustbox{max width=\linewidth}{
      \begin{tabular}{l|c|c}
        \hline \toprule
        $M$  & EPE3D $\downarrow$ & Acc3DS $\uparrow$\\
        \midrule
        3 & 0.040&0.764 \\
        5 & 0.035& 0.806\\
        \cellcolor{gray!20}6&\cellcolor{gray!20}\bf0.025&\cellcolor{gray!20}\bf0.883\\
        8 & 0.031& 0.823\\
        \bottomrule\hline
      \end{tabular}
      }
  \end{subtable}
  \hfill
  \setlength{\tabcolsep}{1.5mm}
  \begin{subtable}[t]{0.31\linewidth}
      \centering
      \caption{Memory Construction.}
      \vspace{-1mm}
      \label{tab:ablation_memory}
      \adjustbox{max width=\linewidth}{
      \begin{tabular}{l|c|c|c|c|c}
        \hline \toprule
        &State & Flow &Uncertainty& EPE3D $\downarrow$ & Acc3DS $\uparrow$ \\
        \midrule
        a)&\ding{55} & \ding{51}& \ding{51} &0.029& 0.834\\
         b)&\ding{51} & \ding{55} & \ding{51}& 0.031& 0.827\\
         c)&\ding{55} & \ding{51} & \ding{55}&0.033 & 0.807 \\
         d)&\ding{51} & \ding{55} & \ding{55}&0.066 &0.415 \\
         e)&\cellcolor{gray!20}\ding{51}& \cellcolor{gray!20}\ding{51} & \cellcolor{gray!20}\ding{51} &\cellcolor{gray!20}\bf 0.025&\cellcolor{gray!20}\bf 0.883 \\
        \bottomrule\hline
      \end{tabular}
      }
  \end{subtable}
\end{table*}

\begin{figure}[t]
    \centering
    \includegraphics[width=1.0\linewidth]{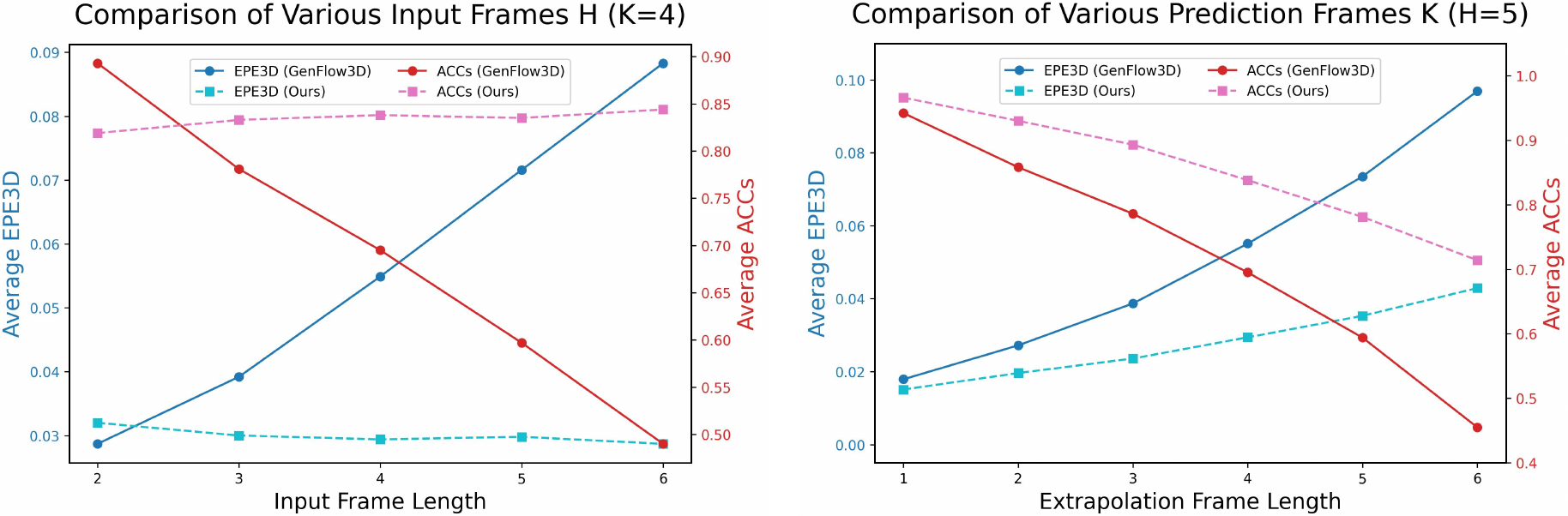}
     \vspace{-0.6cm}
\caption{\textbf{Comparison of different input frames ($H$) and extrapolation frames ($K$) for future scene flow prediction on nuScenes.} 
 }
    \label{fig:visual3}
    \vspace{-0.4cm}
\end{figure}

\section{Discussions}
\textbf{Why MESSENGER Excels Significantly in Prediction?} Our method has significantly superior extrapolation ability as in Table \ref{tab:comparison2} compared to recent SOTA method GenFlow3D \cite{li2025genflow3d}. This advantage can be attributed to three designs: (1) Compared to implicit motion propagation, we explicitly store memory both flow estimates and hidden states, providing strong semantic and motion priors for future prediction. (2) Our designed autoregressive next-frame forecasting paradigm can force the model to unlock temporal dependencies from stored memories. (3) Furthermore, the proposed uncertainty quantification can mitigate long-term drifts commonly existing in autoregressive models. We further demonstrate this by varying the input and extrapolation frame length in Fig. \ref{fig:visual3}. Evaluating the effect of input frames, we freeze the prediction step as $K=4$ and change input frames. From Fig. \ref{fig:visual3} (Left), the baseline method GenFlow3D \cite{li2025genflow3d} witnesses a significant performance drop with more than triple EPE3D increase. In contrast, our method has even improved Acc3Ds and much lower EPE3D. This proves that the baseline method cannot effectively propagate history priors from temporal observations while ours could store and retrieve multiple useful memories, thereby strengthening the future prediction. We also compare different forecasting steps in Fig. \ref{fig:visual3} (Right), where the observed frame length is kept as $H=5$. Both GenFlow3D and our method have a performance drop with longer rollouts due to potential new emerging motions or abrupt dynamics. However, our method holds higher stability with lower metric fluctuations. 

\begin{wrapfigure}{r}{0.45\textwidth}
\centering
\vspace{-0.4cm}
    \includegraphics[width=1.0\linewidth]{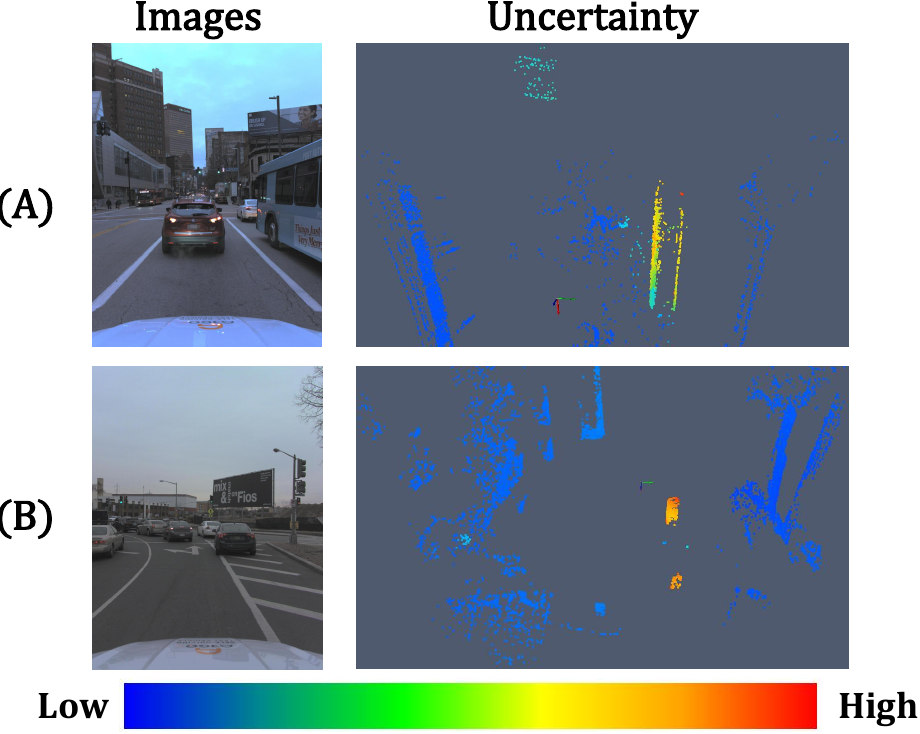}
     \vspace{-0.5cm}
\caption{\textbf{Visualization of uncertainty.} We map the learned uncertainty values to a corresponding color strip. 
 }
    \label{fig:visual4}
    \vspace{-0.4cm}
\end{wrapfigure}

\noindent \textbf{How Uncertainty Works?} We visualize two representative samples with colored uncertainty in Fig. \ref{fig:visual4}, where uncertainty tends to exist in dynamic regions. This is partly because dynamic cars commonly introduce inconsistent motions across frames \cite{liu2023regformer}, where retrieved memory flows may have lower reference reliability to forecast the next frame. In this case, our method will adaptively force the network to learn from current geometric and correlation features, rather than forecasting.

\noindent \textbf{Fine-grained Per-class Error Analysis.} We additionally evaluate the Dynamic Bucket-Normalized EPE metric, proposed in TrackFlow \cite{khatri2024can}, which normalizes per-point EPE by the mean speed within predefined motion buckets and classifies estimates into four categories including regular cars (CAR), other vehicles (OTHER), pedestrians (PED.), and wheeled vulnerable road users (VRU). A direct EPE3D comparison with GenFlow3D on the nuScenes scene flow estimation is shown in Table \ref{tab:per-class1}. Generally, pedestrians and vulnerable road users, such as bicycles or motorcycles, have higher estimation errors than regular cars for both methods. This is partly because PED. and VRU have smaller size, non-rigid motions, and fewer LiDAR points scanning on them, making consistent point correlation across consecutive frames much harder. Our method surpasses GenFlow3D in all dynamic classes, especially in challenging classes PED. and VRU. We also report the Three-way EPE metric \cite{chodosh2024re}, which computes the unweighted average EPE over three categories: foreground dynamic (FD), foreground static (FS) and background static (BS). As in Table \ref{tab:per-class2}, our particular advantage in dynamic regions can be attributed to our designed memory buffer, which can store multi-frame historical flows and features, ensuring stronger motion consistency.

\begin{table*}[t]
  \caption{\textbf{Per-class analysis of estimation errors on nuScenes.}}
  \vspace{-2mm}
  \label{tab:per-class}
  \centering
  \footnotesize
  \setlength{\tabcolsep}{3pt}
  \renewcommand{\arraystretch}{1.2}
  \setlength{\tabcolsep}{1.4mm}
  \begin{subtable}[t]{0.49\linewidth}
      \centering
      \caption{Dynamic Bucket-Normalized EPE.}
      \vspace{-1mm}
      \label{tab:per-class1}
      \adjustbox{max width=\linewidth}{
      \begin{tabular}{l|c|c|c|c}
        \hline\toprule
       $\mathcal{SF}_1$& CAR & OTHER &PED.&VRU \\
        \midrule
       GenFlow3D&0.0234 &0.0257 &0.0353& 0.0418\\
       Ours&\bf0.0209&\bf0.0228&\bf0.0291&\bf0.0321\\
        \bottomrule\hline
      \end{tabular}
      }
  \end{subtable}
  \hfill
  \setlength{\tabcolsep}{1.4mm}
  \begin{subtable}[t]{0.49\linewidth}
      \centering
      \caption{Three-way EPE.}
      \vspace{-1mm}
      \label{tab:per-class2}
      \adjustbox{max width=\linewidth}{
      \begin{tabular}{l|c|c|c|c}
        \hline \toprule
         $\mathcal{SF}_{H+4}$  & FD & FS&BS&Class Average\\
        \midrule
         GenFlow3D&0.137 &0.072 &0.086& 0.098\\
       Ours&\bf0.051&\bf0.013&\bf0.024&\bf0.029\\
        \bottomrule\hline
      \end{tabular}
      }
  \end{subtable}
  \vspace{-3mm}
\end{table*}

\section{Conclusion}
In this paper, we proposed a novel memory-enhanced sequential scene flow estimation pipeline based on autoregressive next-frame forecasting. The core design is a dynamically maintained memory buffer which stores explicit coarse flows, implicit states, and reliability scores from multiple history timestamps. The current flow is initialized by progressively forecasting the next frame motions based on retrieved memory features. To alleviate potential compounding errors and filter unreliable outliers, we develop an uncertainty measurement to indicate per-point trustworthy score when weighting history memories. Extensive experiments demonstrate our method outperforms existing scene flow methods in both estimation and prediction, especially in long-horizon future predictions.

\textbf{Limitations and Future Work.} Although our work based on attention-based autoregressive forecasting has future prediction capability within a limited frame length, future work could investigate how to incorporate large-scale pretrained backbone such as Wan \cite{wan2025wan} for long-horizon world modeling \cite{jiang2026vectorworld,liu2026towards,dong2026towards,dong2026havln20openbenchmark,liu2026driveva,lu2026worldcoder}. Scene flow represents dense 4D world dynamics, which can strengthen motion modeling in recent world models.

\textbf{Acknowledgments.} We thank Qingwen Zhang from KTH for her assistance in running open-source codes for efficiency comparison. In addition, Jiuming Liu appreciates the Trinity Hall and Cambrdige Trust for the International Scholarship they provided.
{
\small
\bibliographystyle{unsrt}
\bibliography{main}
}






\end{document}